\documentclass[runningheads]{llncs}
\usepackage[T1]{fontenc}
\usepackage{graphicx}
\usepackage{xcolor} % permite destacar o texto por cor
\usepackage{booktabs}
\usepackage{amsmath} % formulas matematicas
\usepackage{amssymb}

\newif\ifanonym
 \anonymfalse  % versão com autor
\begin{document}
\title{Memory-Aware Multi-Agent Architecture with LLM-based Orchestration for Conversational Recommendation}
\titlerunning{Memory-Aware Multi-Agent Architecture for CRS}

\ifanonym
    \author{Anonymous Authors}
    \authorrunning{Anonymous}
\else
    \author{
% ICTi
Emilly Pitman de Castro\inst{1}, Vanessa Silva\inst{1}, Gustavo Kanno\inst{1}, Marcus Santos\inst{1}, 
Bruno Ferraz do Amaral\inst{1}, Marcelo Nogueira\inst{1}, Lucas Francisco Pellicer\inst{1}
\\[0.4em]
% Eldorado
Lucas Lanaro\inst{2}, Pedro Augusto Luiz\inst{2}, João Renato Ribeiro Manesco\inst{2}, 
Giovanni Carbinatti\inst{2}, Isabela de Lima Santos\inst{2}, Lucas Oliveira Jeronymo\inst{2}, 
Jeison Fappi\inst{2}, Raul Rotilli Aguirre\inst{2}, Fabio André Frantz\inst{2}, 
Renata De Paris\inst{2}, Fabiana Marucci\inst{2}, Lohan Silva\inst{2}, 
Vinicius Diniz Reolon\inst{2}, Wandemberg Gibaut\inst{2}
}
\institute{
Instituto de Ciência e Tecnologia Itaú (ICTi), Brasil\\
{\footnotesize\texttt{
emilly.castro, vanessa-fernandes.silva, gustavo.kanno,
marcus.cognolatto-santos,\\
bruno.ferraz-amaral, marcelo.b.nogueira,
lucas.pellicer@itau-unibanco.com.br
}}
\and
Instituto de Pesquisas Eldorado, Brasil\\
{\footnotesize\texttt{
lucas.lanaro, pedro.augusto, joao.manesco.BE,
giovanni.carbinatti,\\
isabela.santos, lucas.jeronymo, jeison.fappi,
raul.aguirre, fabio.frantz,\\
renata.paris, fabiana.marucci, lohan.silva,
vinicius.reolon, wandemberg.gibaut@eldorado.org.br
}}
}

%\author{Anonymous Authors}
%\authorrunning{Anonymous Authors}
%\institute{Anonymous Institution}
%author{Lucas Lanaro\inst{1}\orcidID{0000-1111-2222-3333} \and
%Second Author\inst{2,3}\orcidID{1111-2222-3333-4444} \and
%Third Author\inst{3}\orcidID{2222--3333-4444-5555}}
%
    \authorrunning{Castro et al.}
\fi
%\institute{Princeton University, Princeton NJ 08544, USA \and
%Springer Heidelberg, Tiergartenstr. 17, 69121 Heidelberg, Germany
\maketitle              % typeset the header of the contribution
\begin{abstract} % 232 words
Conversational recommender systems often treat user sessions in isolation, limiting their ability to capture how user intent evolves over time. While large language models (LLMs) have improved semantic understanding, they are typically used as isolated components, without addressing how memory, retrieval, and decision-making should be integrated at the system level.
In this work, we propose a memory-aware multi-agent architecture for conversational recommendation, in which LLMs act as orchestration layers that coordinate intent refinement, query construction, and interaction flow. A key design principle is the integration of long-term memory as a first-class component, influencing both routing decisions and semantic query formation.
We evaluate the system through cross-session memory experiments, multi-turn conversational simulations, and supervisor-driven interaction studies on the Amazon Reviews dataset. Results show that memory-enabled interactions recover up to 63\% of the gap toward explicit-query scenarios and prevent semantic degradation in extended dialogues. In addition, supervisory orchestration increases user acceptance by +28.9 percentage points, highlighting the importance of structured orchestration in guiding interaction and output quality.
We also observe an exposure bottleneck between retrieved candidates and displayed recommendations. While LLM-reranking does not improve Hit@K, it consistently enhances semantic consistency and diversity. To capture these effects, we introduce the Semantic Consistency Score (SCS), a complementary evaluation measure. Overall, the results emphasize the importance of system-level design choices, suggesting that effective conversational recommendation benefits from explicit memory integration, orchestration, and evaluation beyond traditional ranking metrics.
%The abstract should briefly summarize the contents of the paper in 150--250 words.

%{\color{red}Resumo bem condensado do artigo inteiro. De 150 a 250 palavras.

%Estrutura:
%Qual problema estamos tentando resolver ($25 - 30\%$)?
%Qual abordagem usamos ($25 - 30\%$)? 
%Como avaliamos ($15 - 20\%$)? 
%Qual o principal resultado/contribuição ($30\%$)?}

%\keywords{Recommender Systems  \and Multi-Agent Systems \and User Modeling \and Large Language Models \and Synthetic Personas \and Personalization.}
\end{abstract}
\section{Introduction}

Conversational recommender systems aim to iteratively infer user preferences through interaction, as intent is rarely explicit and instead evolves across dialogue turns. However, most existing approaches treat sessions in isolation, discarding preference signals once interactions end, which prevents systems from leveraging accumulated user context across interactions. This limits their ability to model long-term intent and maintain coherence over extended dialogues \cite{jannach2021survey,jannach2023evaluating}.

Recent advances in large language models (LLMs) have enabled new approaches to conversational recommendation, including multi-agent architectures where components specialize in tasks such as intent understanding, retrieval, and response generation \cite{hou2024bridging}. While effective for intent interpretation, these approaches typically operate as loosely integrated components, without a unified system-level design that governs how information flows across interaction, retrieval, and ranking stages \cite{jannach2023evaluating}. As a result, there is limited understanding of how these components interact in multi-turn and cross-session scenarios, particularly in architectures where LLMs act as orchestration layers.

This work presents a memory-aware multi-agent architecture in which an LLM-based supervisor coordinates interaction, integrates long-term memory into query construction, and guides recommendation decisions across turns.

The described approach is based on a multi-agent framework in which an LLM-based supervisor coordinates interactions, integrates long-term memory into query construction, and guides recommendation decisions across turns. The architecture explicitly incorporates cross-session memory as a first-class component, enabling the system to leverage past interactions when the user’s current intent is not clearly specified.

Our main contributions are:

\begin{itemize}
    \item A memory-aware multi-agent architecture for conversational recommendation with LLM-based orchestration.
    \item An empirical analysis of the impact of orchestration and cross-session memory on recommendation effectiveness and interaction dynamics in multi-turn settings.
    \item The identification of a gap between retrieval and exposure in conversational recommendation.
    \item The Semantic Consistency Score (SCS) as a complementary metric for evaluating semantic alignment.
\end{itemize}

Experimental results indicate that memory reduces a substantial portion of the performance gap in underspecified scenarios, while conversational orchestration leads to higher acceptance rates. Additionally, LLM-based reranking enhances semantic alignment and diversity, although not always improving traditional ranking metrics. These findings highlight the importance of considering system-level interactions when designing conversational recommender systems.

% Rather than focusing on isolated improvements, we analyze how system-level design choices—particularly memory integration and orchestration—affect recommendation behavior. Experimental results show that memory improves performance in underspecified scenarios, recovering up to 63\% of the gap toward explicit-query baselines, while conversational orchestration increases user acceptance by +28.9 percentage points. LLM-based reranking enhances semantic consistency and diversity, with limited gains in traditional ranking metrics.

% Our analysis further reveals a structural gap between retrieval and exposure, where relevant items are identified but not consistently surfaced to users. These findings indicate that conversational recommendation performance depends not only on modeling user intent, but also on how system components interact to support effective decision-making.

\section{Related Works}
\label{sec:relatedworks}

Conversational recommender systems (CRS) iteratively refine user preferences through dialogue~\cite{jannach2021survey,zhang2018towards}. However, most approaches operate at the session level, without persistent memory across interactions, limiting their ability to model long-term user intent~\cite{jannach2023evaluating}.

Large Language Models (LLMs) have been increasingly adopted in recommendation pipelines, for example as zero-shot rankers or conversational agents~\cite{hou2024large,feng2023large}. While these approaches improve flexibility and semantic understanding, they typically treat LLMs as standalone components and do not address system-level integration with retrieval and memory.

Recent work on LLM-based agents emphasizes memory for long-term reasoning and adaptation, including approaches such as Generative Agents~\cite{park2023generative}, MemoryBank~\cite{zhong2024memorybank}, and Reflexion~\cite{shinn2023reflexion}. However, these mechanisms are not directly designed for recommendation pipelines, where memory must influence retrieval and ranking decisions.

In parallel, LLMs have been explored as coordination layers for multi-step reasoning and tool use~\cite{schick2023toolformer}, highlighting their potential for orchestrating complex pipelines. However, prior work typically studies memory, retrieval, ranking, or orchestration as separate mechanisms, without analyzing how cross-session memory should act as a first-class signal that influences query construction, routing, and recommendation decisions across turns. To the best of our knowledge, no prior work jointly investigates long-term memory, LLM-based orchestration, and hybrid retrieval-reranking interactions at the system level for conversational recommendation. Our work addresses this gap through a unified architecture and a controlled empirical analysis of how these components shape multi-turn recommendation behavior.

\section{Architecture}
\label{sec:architecture}

\subsection{System Architecture}
We propose a multi-agent architecture that integrates LLM-based agents with classical recommendation algorithms, orchestrated through a structured pipeline. Each agent fulfills a specific role, such as intent refinement, routing, or recommendation, while sharing access to memory and retrieval components. This design enforces a clear separation of responsibilities and enables consistent behavior across multi-turn interactions. 

\begin{figure}[h]
    \centering
    \includegraphics[width=1\linewidth]{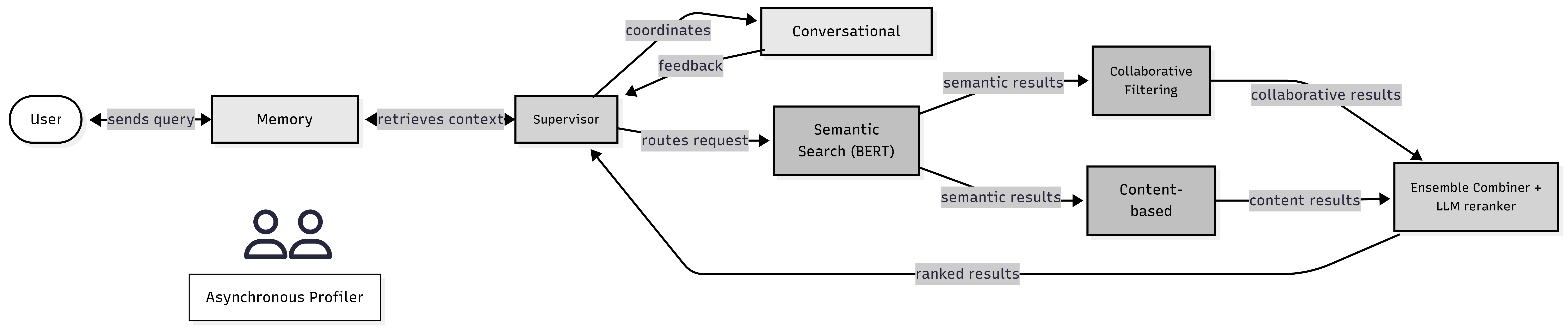}
    \caption{Overall system architecture and interaction flow of the proposed memory-aware multi-agent architecture.}
    \label{fig:diagrama_geral}
\end{figure}

As illustrated in Figure~\ref{fig:diagrama_geral},a Supervisor Agent acts as the central orchestrator between the user and internal system components. It maintains control over the interaction flow across stages including intent refinement, query construction, routing, and response generation, coordinating the transition between preference elicitation and recommendation. Memory is accessed prior to orchestration, enriching the user query with cross-session context before routing decisions are made. The architecture is implemented using LangGraph, enabling explicit control over execution flow and state management across agents, consistent with LLM-based coordination approaches~\cite{schick2023toolformer}.

\subsection{User Modeling and Interaction }

\subsubsection{Supervisory Agent}

The Supervisor acts as the central coordination layer, responsible for managing interaction flow and refining user queries based on conversational and memory context. It integrates information from the current dialogue and memory retrieval to guide downstream components.

Rather than operating as a generic router, it structures the interaction by eliciting missing preferences and updating query representations across turns, enabling the system to progressively align recommendations with user intent, particularly in underspecified scenarios.

In this role, it coordinates interaction and query refinement while guiding the behavior of downstream components. Its design is inspired by recent work on LLM-based control and self-improvement mechanisms\cite{schick2023toolformer,shinn2023reflexion}.

The decision process is driven by the interaction state, retrieved candidates, and dialogue history, allowing the system to balance preference elicitation and recommendation generation across turns.

\subsubsection{Conversational Agent}
% Despite its name, the Conversational Agent is not a general-purpose dialogue agent \revisao{acho que podem referenciar a Figure \ref{fig:diagrama_geral} aqui}; rather, it serves as a specialized one, with the main objective of progressively mapping the user's preference landscape by creating targeted questions. The agent models the user's needs across multiple dimensions, including functional requirements, usage context, and brand preferences, among others, and generates follow-up questions that narrow the space of plausible recommendations. For instance, upon learning that a user seeks headphones for professional use, the agent may probe further into noise-cancellation preferences or connectivity requirements.

% Beyond refinement, the agent also handles product-related questions, out-of-scope queries are declined, and the user is redirected to the recommendation task. To support extended sessions, the agent maintains two complementary memory mechanisms: a short-term context window comprising the current conversation history and a long-term memory module that enables retrieval of user preferences from prior sessions. %In architectural terms, the agent is implemented as a graph node conditioned on a language model operating under structured instruction and constraint sets. % Acho que isso já fica implicito na figura.

The Conversational Agent is responsible for elicitating user preferences through targeted follow-up questions. As shown in Figure~\ref{fig:diagrama_geral}, it operates under the Supervisor, generating structured interactions that reduce ambiguity in user intent and narrow the candidate space.

Its behaviour is guided by the current conversation state and retrieved memory context, enabling efficient preference refinement across turns~\cite{park2023generative}

\subsection{Recommendation Engine}
The recommendation engine, whose processing flow is illustrated in Figure~\ref{fig:diagrama_geral}, is implemented as a modular pipeline combining multiple retrieval and ranking strategies. It integrates:

\textbf{Semantic Retrieval} Dense representations from Sentence-BERT are used for similarity search between natural-language queries and items, as they capture semantic intent beyond lexical overlap \cite{reimers2019sentence}. Item embeddings are indexed with FAISS to support efficient approximate nearest-neighbor retrieval over the catalog \cite{johnson2019billion}. We apply Maximal Marginal Relevance (MMR) to balance relevance and diversity, reducing redundancy in candidates retrieved by semantic similarity alone \cite{carbonell1998use}.

\textbf{Collaborative Filtering} We adopt Alternating Least Squares (ALS), a well-established implicit-feedback matrix factorization method, to capture latent user--item interaction patterns that complement textual and semantic signals \cite{hu2008collaborative}.

\textbf{Content-Based Filtering} A content-based module uses TF-IDF representations extracted from textual features such as product descriptions and reviews, providing a lexical signal complementary to dense semantic retrieval \cite{manning2008introduction}.

\textbf{Ensemble Strategy and LLM-based Reranking} Results from the retrieval components are combined through a weighted ensemble, allowing heterogeneous collaborative, lexical, and semantic signals to contribute to candidate generation \cite{ricci2022recommender}. The candidate set is then refined by an LLM-based reranker, which operates only over retrieved items and prioritizes semantic relevance and contextual coherence rather than expanding the retrieval space \cite{hou2024large}.
  
\subsection{Memory Orchestrator}

The memory module provides cross-session context by storing and retrieving user-related information accumulated over previous interactions. Memory is accessed prior to orchestration, allowing retrieved signals to directly influence query construction and routing decisions.

Information is organized into structured nodes representing user preferences, constraints, and contextual signals. Updates follow an append-only strategy, where new nodes reference previous state for the same subject, forming a linked sequence of states over time while maintaining a single active representation.

This design allows the system to preserve and reuse user context across interactions, supporting more consistent recommendations in underspecified scenarios. It is conceptually aligned with recent approaches to memory-augmented LLM systems \cite{zhong2024memorybank,shinn2023reflexion}, where external memory is used to extend context beyond individual interactions.

However, as user preferences evolve, previously stored signals may introduce noise, indicating that memory effectiveness depends on the stability of user intent.

\section{Experimental Methodology}
\label{sec:methodology}

\subsection{Dataset and Preprocessing}

Experiments are conducted on the Amazon Reviews 2023 Electronics dataset \cite{ni2019justifying,mcauley2023amazon}, which provides both structured interaction data and rich textual content, enabling query-driven and conversational recommendation scenarios.

To support controlled evaluation of LLM-intensive components, a subset of the 1,000 most active users is constructed (January 2004–May 2023), yielding 19,678 unique items and 54,521 interactions (sparsity: 99.72\%), with a median of 54 interactions per user. Each record includes \texttt{user\_id}, \texttt{ASIN}, timestamps, ratings, review text, and product metadata. Recommendation and evaluation rely on a \textit{product-centric view}, focusing on items with available product metadata.

This subset is chosen to enable controlled analysis of architectural effects under computationally intensive settings, including multi-turn interactions and cross-session memory. While limited in scale, it allows isolating the impact of design choices such as orchestration and memory integration.

Experiments are conducted in a controlled environment to enable systematic comparison of components, rather than providing absolute performance estimates.

Given the limited scale of the resulting subset, results should be interpreted mainly through relative comparisons across configurations, as statistical power for ranking metrics such as Hit@K \cite{tamm2021quality} and NDCG@K \cite{jarvelin2002cumulated} is constrained and collaborative filtering effectiveness is reduced by sparse user neighborhoods.

\subsection{Offline Evaluation of the Recommendation Pipeline}
We adopt a Leave-Last-3-Out protocol with temporal consistency and synthetically generated queries from review text. Five models of increasing complexity are evaluated: Random, Popularity, Semantic retrieval (Sentence-BERT), Prototype (semantic + ALS + content-based), and Prototype + LLM (LLM-based reranking). These configurations cover standard recommender system families, while the hybrid prototype combines ALS-based collaborative filtering with sparse TF-IDF and dense Sentence-BERT retrieval \cite{ricci2022recommender,hu2008collaborative}.

Ranking performance is measured using Hit@K and NDCG@K, with intra-list diversity assessed through ILS@K following Ziegler et al.~\cite{ziegler2005topicdiversification}. To complement these metrics in terms of semantic alignment under sparse settings~\cite{krichene2020sampled}, we introduce the Semantic Consistency Score (SCS), an LLM-based evaluation signal that assesses whether recommended items are semantically consistent with the inferred user intent.

Formally, for a user $u$ with query $q_u$ and recommended list $\mathcal{R}_u=\{r_1,\dots,r_K\}$, an LLM evaluator returns a binary judgment $c_i \in \{0,1\}$ per item, where $c_i=1$ if $r_i$ satisfies the constraints expressed in $q_u$. The per-user and global scores are defined as:

\begin{equation*}
\text{SCS}_u = \frac{1}{K}\sum_{i=1}^{K} c_i,\quad
\text{SCS} = \frac{1}{|\mathcal{U}|}\sum_{u \in \mathcal{U}} \text{SCS}_u
\end{equation*}
%% [23/06/26] Lanaro veja se assim ja resolve:
where $K$ denotes the recommendation list size and $\mathcal{U}$ denotes the evaluated user set, with $|\mathcal{U}|$ representing the number of users.

The evaluator operates with temperature 0.0 for deterministic judgments. Unclear classifications are treated as consistent to avoid penalizing ambiguous cases, which may slightly bias scores upward. As an LLM-based metric, SCS should therefore be interpreted as a complementary signal rather than a standalone evaluation measure.

\subsection{Intent-based Conversational Evaluation}

Conversational intent recovery is evaluated on 998 users with at least six interactions. For each user, one interacted item is held out as the target and removed from the history. A Claude 3 Haiku proxy user seeks the held-out item without naming it, progressively refining the request for up to six turns. No temporal split is enforced, as the goal is intent recovery rather than next-item prediction.

Cumulative \emph{Hit@T} extends the standard Hit@K metric to multi-turn recommendation by measuring whether the target item has been retrieved at least once up to interaction turn $T$. Together with \emph{MRR} (Mean Reciprocal Rank) \cite{tamm2021quality} computed over the accumulated ranked lists across turns, it serves as the primary evaluation metric. Results are reported under two views: \emph{internal} (target retrieved by the pipeline) and \emph{shown} (target displayed to the user).

\subsection{Supervisor Evaluation}

A three-tier ablation is conducted: \textbf{Baseline} issues recommendations from a single-shot query; \textbf{Conversational} adds a supervisor capable of requesting clarification; \textbf{Conv.\ w/\ Memory} further activates the Memory Orchestrator to persist preferences across turns. All tiers share the same retrieval engine and evaluation partition.

Following established simulation practice~\cite{jannach2023evaluating,fang2024multi}, a proxy user is initialized with the category, attributes, and price of a held-out target but may not name it, disclosing requirements organically~\cite{zhu2024reliable}. The engine retrieves 10 items; the top 3 are presented. An item is accepted if it matches category and features, and its price falls within 30\% of the target.

Performance is assessed along two dimensions. Recommendation quality is captured by \textbf{Hit Rate} (target in top-10 retrieved), \textbf{Acceptance Rate} (any item satisfying the latent-match rule), \textbf{Lift} (absolute acceptance gain over Baseline), and \textbf{Conversions} (sessions where an initial rejection recovers to acceptance). Dialogue efficiency is measured by \textbf{Average Turns} and \textbf{Recommendation Latency} (mean turn at which the first ranked list is issued).

\subsection{Cross Session Memory Evaluation}

To evaluate the effect of cross-session long-term memory on recommendation quality, we conducted a two-session experiment using 30 synthetic user profiles with gradually revealed preferences. In Session~1, a multi-turn dialogue was simulated and the extracted preferences were stored in a long-term memory module. In Session~2, the same vague query was issued under two conditions: Condition~A (without memory) and Condition~B (with memory from Session 1 restored), with all other system components fixed. Recommendation quality was evaluated using a deterministic LLM-as-a-judge protocol~\cite{zheng2023judging}, which performed pairwise comparisons between conditions based on the user’s revealed preferences and Session~1 dialogue, assigning an alignment score (0--10) per list and a winner. We report win rate (B over A) and mean alignment score per condition.
\section{Results}
\label{sec:results}

\subsection{Offline Evaluation}

\begin{figure}[htbp]
\begin{center}
\includegraphics[width=1.0\linewidth]{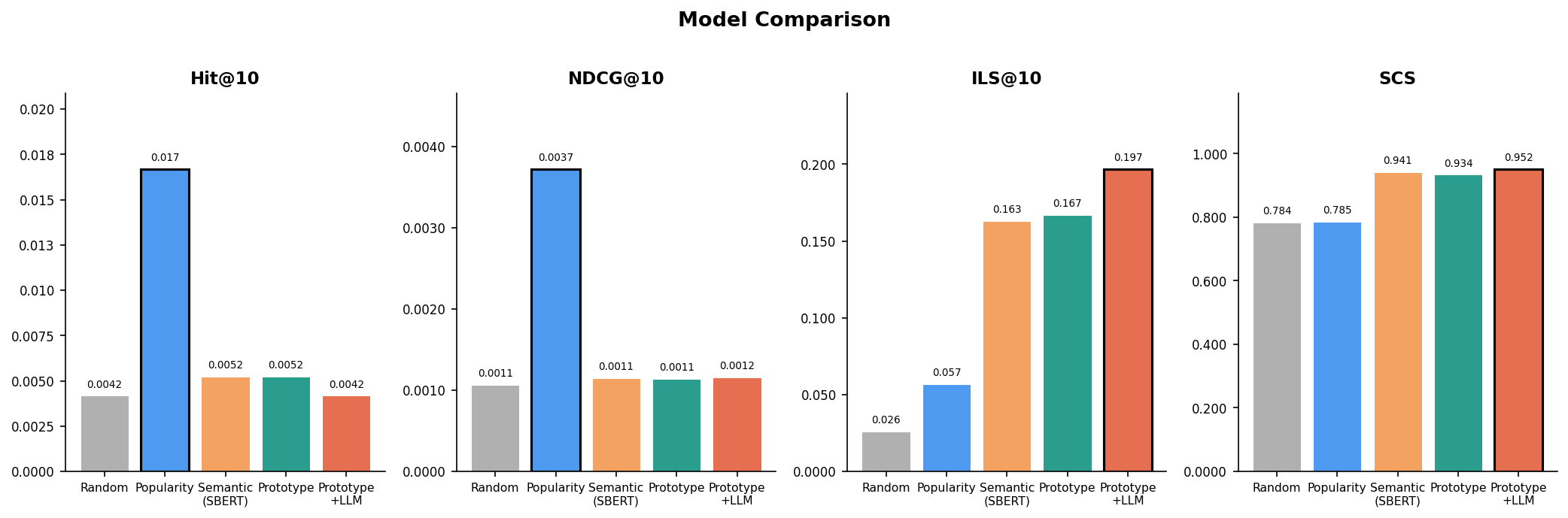}
\caption{Performance comparison across pipeline stages.}
\label{fig:model_comparison}
\end{center}
\end{figure}

Figure~\ref{fig:model_comparison} reports Hit@10, NDCG@10, ILS@10, and Semantic Consistency Score (SCS) across all pipeline stages.The popularity baseline achieves the highest ranking accuracy (Hit@10 = 0.0167, NDCG@10 = 0.0037), reflecting its bias toward frequently interacted items, but yields lower semantic consistency (SCS = 0.785).

Semantic and hybrid approaches trade ranking accuracy for diversity and intent alignment. The proposed hybrid architecture improves both diversity (ILS@10 = 0.167) and semantic consistency (SCS = 0.930) over Semantic and Popularity baselines. Adding LLM-based reranking further increases diversity (ILS@10 = 0.197) and semantic consistency (SCS = 0.954), with a slight reduction in ranking accuracy, highlighting a trade-off between exact-match retrieval and semantic alignment.

\begin{figure}[h]
\centering
\includegraphics[width=0.5\linewidth]{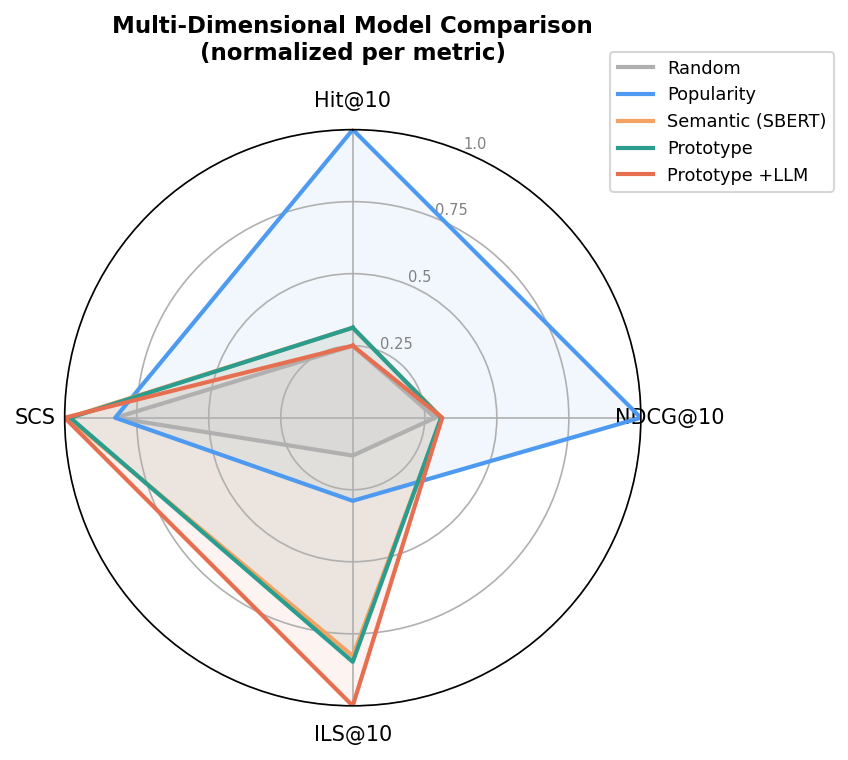}
\caption{Multi-dimensional model comparison (normalized metrics).}
\label{fig:radar}
\end{figure}

Figure~\ref{fig:radar} provides a multi-dimensional comparison highlighting the trade-offs between accuracy, diversity, and semantic consistency across models.

\subsubsection{Ablation Study} 
\begin{figure}[htbp]
\centering
\includegraphics[width=0.8\textwidth]{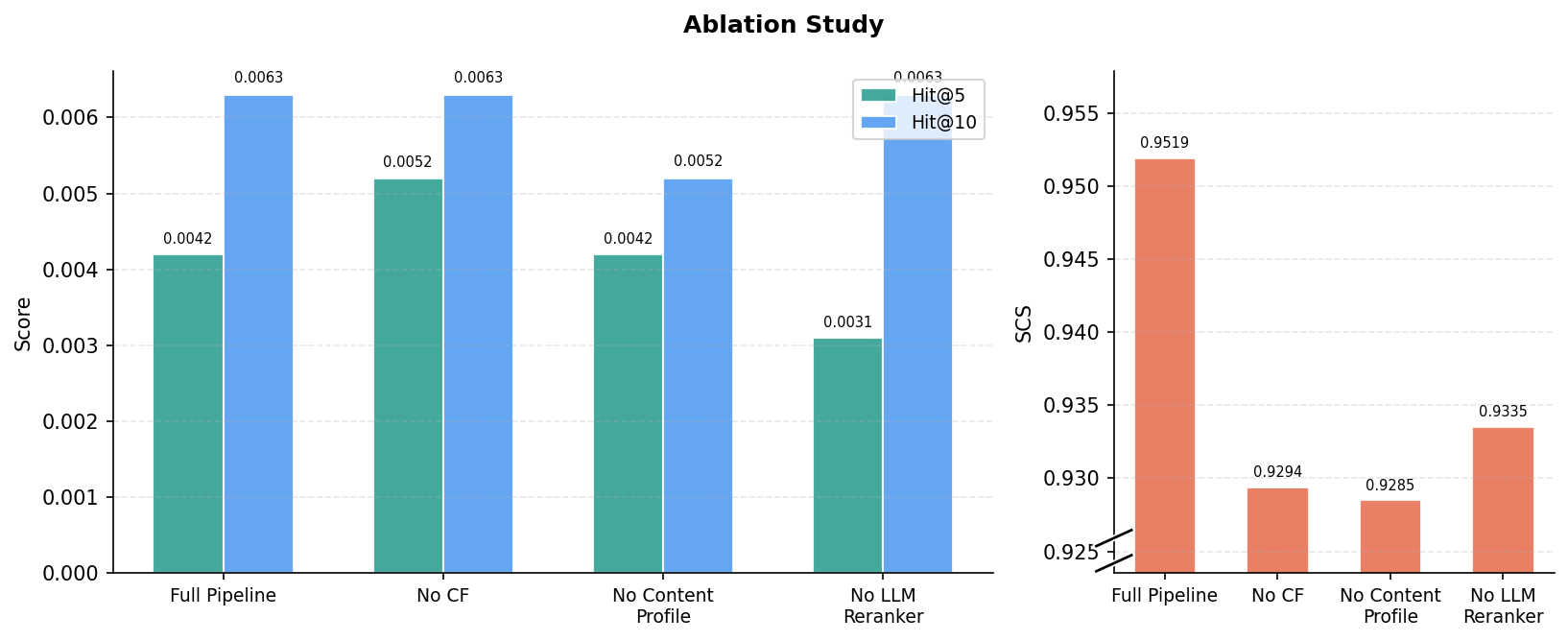}
\caption{Impact of component removal on pipeline performance.}
\label{fig:ablation}
\end{figure}

The ablation results are presented in Figure~\ref{fig:ablation}.

Removing the content-based component leads to the largest drop in ranking performance, particularly in Hit@10 and NDCG@10, indicating its central role in item relevance. Removing collaborative filtering has limited impact, suggesting a complementary contribution. Disabling the LLM reranker produces minimal changes in Hit@10 but reduces semantic consistency, confirming its role in improving alignment rather than retrieval accuracy.

\subsection{Intent-based evaluation}

Table~\ref{tab:intent_hit_at_t} reports cumulative \emph{Hit@T} under both evaluation views. By turn 6, the target appears in 74.75\% of sessions under the internal view and 62.63\% under the shown view, with MRR reaching 0.2836 and 0.2954, respectively.

Performance improves rapidly in early turns and then plateaus, indicating that most intent is captured within the first interactions. The gap between internal and shown views reveals an exposure bottleneck, where relevant items are retrieved but not surfaced to the user.

\begin{table}[t]
\centering
\small
\begin{tabular}{c c c}
\hline
$T$ & Internal Hit@T & Shown Hit@T \\
\hline
1 & 0.4639 & 0.2836 \\
2 & 0.6303 & 0.4459 \\
3 & 0.6804 & 0.5331 \\
4 & 0.7124 & 0.5731 \\
5 & 0.7295 & 0.6022 \\
6 & 0.7475 & 0.6263 \\
\hline
\end{tabular}
\caption{Cumulative Hit@T under the internal and shown evaluation views.}
\label{tab:intent_hit_at_t}
\end{table}

\subsection{Supervisor Evaluation}

Table~\ref{tab:ablation_sup} presents the ablation study across three configurations. The conversational tier recovers 25.3\% of Baseline failures into acceptances through dialogue alone, raising overall Acceptance Rate from 37.6\% to 57.9\% (+20.3~pp). Adding long-term memory further increases acceptance to 66.5\% (+28.9~pp), with 31.2\% of Baseline failures recovered, suggesting that cross-session context compounds the gains of in-session clarification in the evaluated setting.

Notably, Acceptance Rate (66.5\%) exceeds Hit Rate (51.6\%) in Conv.\ w/\ Memory, indicating that the reranker, guided by conversational context, frequently surfaces adequate alternatives even when the exact target is never retrieved.

\begin{table}[ht!]
\caption{Ablation study comparing three system configurations. All values in percentage points. Lift is relative to Baseline.}
\label{tab:ablation_sup}
\centering
\small
\setlength{\tabcolsep}{3pt}
\begin{tabular}{@{}lcccc@{}}
\toprule
& \textbf{Hit Rate} & \textbf{Acceptance} & \textbf{Lift} & \textbf{Conversions} \\
\midrule
Baseline          & 50.7 & 37.6 & {---}   & {---} \\
Conversational    & 51.1 & 57.9 & {+20.3} & 25.3  \\
Conv.\ w/\ Memory & 51.6 & \textbf{66.5} & \textbf{+28.9} & \textbf{31.2} \\
\bottomrule
\end{tabular}
\end{table}

Efficiency metrics (Table~\ref{tab:orchestrator}) show that successful sessions converge quickly (1.44 and 1.71 turns), while failures require longer interactions (2.98 vs. 7.59), reflecting early commitment and subsequent recovery behavior

\begin{table}[ht!]
\caption{Orchestrator efficiency metrics conditioned on session outcome. Rec.\ Latency is the average turn at which the first recommendation is issued.}
\label{tab:orchestrator}
\centering
\small
\setlength{\tabcolsep}{2pt}
\begin{tabular}{@{}lcccc@{}}
\toprule
& \textbf{Turns (Overall)} & \textbf{Turns (Success)} & \textbf{Turns (Failure)} & \textbf{Rec.\ Latency} \\
\midrule
Baseline           & 1.00 & 1.00 & 1.00 & 1.00 \\
Conversational     & 2.09 & 1.44 & 2.98 & 1.16 \\
Conv.\ w/ Memory    & 3.68 & 1.71 & 7.59 & 1.02 \\
\bottomrule
\end{tabular}
\end{table}

\subsection{Cross Session Memory Evaluation}

% We evaluate the effect of cross-session long-term memory on recommendation quality under underspecified user input using a fully simulated experimental pipeline. To this end, 30 synthetic user profiles are generated via an LLM, spanning diverse product categories, budgets, and preference structures, with instructions to reveal preferences gradually. For each profile, Session 1 is simulated as a multi-turn dialogue, during which stated preferences are stored in a long-term memory module. In Session 2, the same vague query is issued under two conditions: A, with the memory store cleared, and B, with memory from Session 1 restored, while all other system components remain fixed. Recommendation quality is assessed using a blind LLM judge, which compares the generated lists against preferences revealed in Session 1. As shown in Figure \ref{fig:cross_session}, across the profiles, recommendations with memory achieve higher alignment ($\mu = 5.60$) than the no-memory baseline ($\mu = 4.07$). In pairwise comparisons, the memory condition wins in 57\% of cases, the no-memory baseline in 33\%, with 10\% ties, indicating that cross-session memory consistently improves preference recovery under vague queries. Limitations include reliance on LLM-based simulation and evaluation, and restriction to short-term transfer.

\begin{figure}[h!]
\centering
\includegraphics[width=0.75\textwidth]{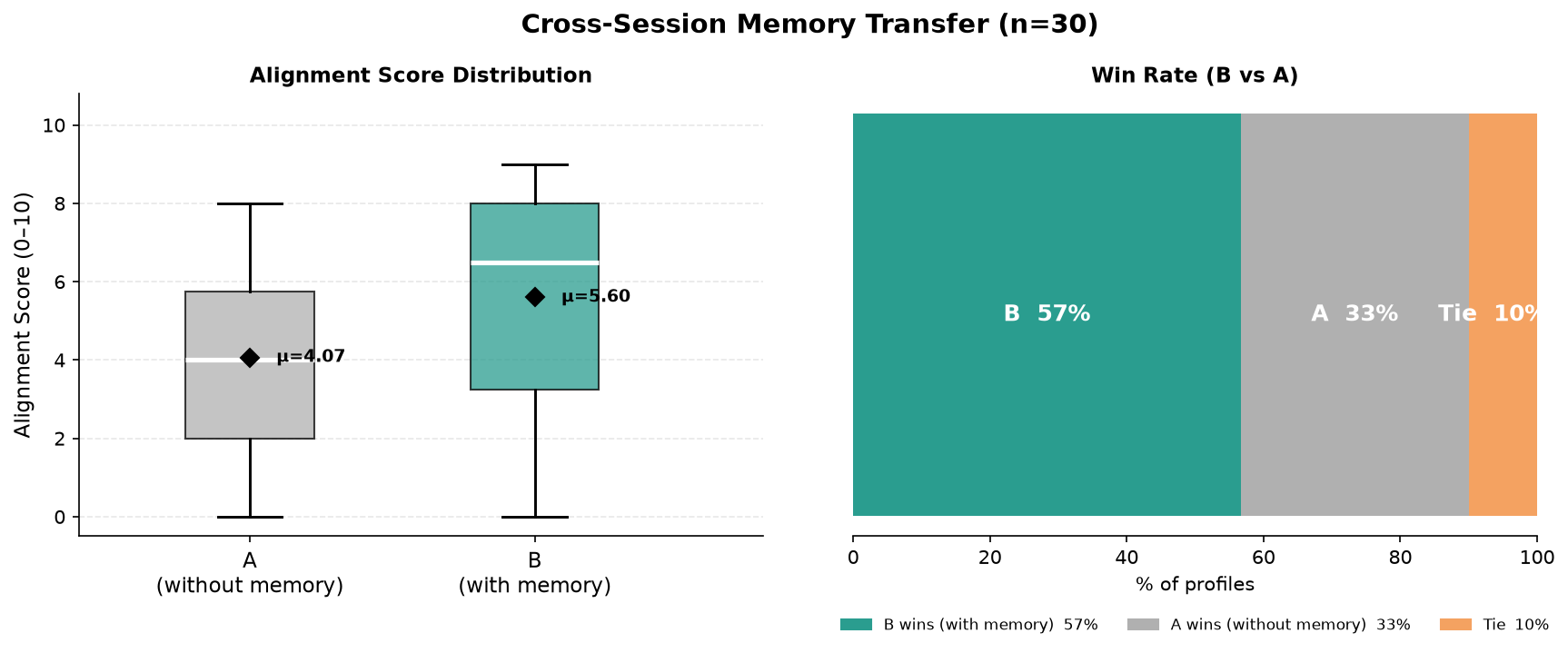}
\caption{Impact of Long Term Memory in Cross Session.}
\label{fig:cross_session}
\end{figure}

The inclusion of cross-session memory indicated a consistent improvement in preference recovery when handling vague queries. As illustrated in Figure \ref{fig:cross_session}, recommendations generated under Condition B (with memory restored) achieved a higher average alignment score ($\mu=5.60$) compared to the baseline Condition A ($\mu=4.07$).

Furthermore, pairwise comparisons between the two conditions highlight the robust performance of the memory-augmented system. Condition B outperformed the baseline in 57\% of the evaluated cases, while Condition A was preferred in only 33\% of cases, with the remaining 10\% resulting in ties. These findings indicate that integrating cross-session memory consistently enhances the system's ability to tailor recommendations to implicit user needs.

% It is important to note however, certain limitations of this experiment, primarily its reliance on LLM-based simulation and evaluation, as well as its restriction to memory derived from a single prior conversation, rather than a series of interactions expected of true long-term memory.

\section{Discussions}
\label{sec:discuss}

\subsection{Offline and Intent-based Evaluations}

Popularity-based methods achieve higher Hit@K and NDCG@K but exhibit lower semantic consistency, reflecting their sensitivity to interaction frequency rather than user intent, a well-documented effect of popularity bias~\cite{abdollahpouri2019managing}.

To address this, the Semantic Consistency Score (SCS) complements traditional ranking metrics by measuring alignment between recommendations and inferred user intent. While standard metrics emphasize exact matches and interaction frequency, SCS provides a semantic perspective that is particularly relevant in conversational settings, where user preferences are often underspecified. In our experiments, this metric reveals differences across methods that are not reflected in ranking-based evaluation alone.

Semantic and hybrid approaches improve diversity and alignment, indicating that representation-based retrieval better captures intent beyond historical interactions~\cite{reimers2019sentence,hou2024bridging}. LLM-based reranking follows prior observations on LLM rankers by improving semantic coherence despite limited Hit@K gains~\cite{hou2024large}. In our setting, this trade-off extends to conversational recommendation, where reranking interacts with orchestration and memory and is most appropriate when semantic alignment and diversity justify its computational cost.

Intent-based evaluation provides additional insight into interaction dynamics. Most user intent is recovered within the first turns before stabilizing. A consistent gap between retrieval and exposure indicates that recommendation quality depends not only on retrieval effectiveness, but also on how candidates are filtered, ranked, and presented during interaction. 

These results should be interpreted in the context of the evaluation setting, where ground truth reflects historical behavior rather than the full space of user preferences. As a result, optimizing for exact matches may not align with capturing evolving intent, highlighting a central trade-off in conversational recommendation. Given the controlled and relatively small-scale setup, performance differences should be interpreted as indicative trends rather than statistically robust estimates. 

\subsection{Conversational Orchestration}
The supervisor has a clear impact on interaction quality. Introducing a clarification-capable supervisor increases acceptance rates by up to 28.9 percentage points, enabling structured preference elicitation under underspecified intent.

Orchestration shapes exploration across turns by iteratively refining user preferences and updating query representations, allowing progressive alignment with intent rather than relying on a single-step prediction. These effects are reinforced by LLM-based reranking, which leverages accumulated conversational context to surface relevant alternatives. 

Efficiency results reveal a trade-off: successful sessions converge quickly, while failures require longer interactions. This reflects early commitment when intent is clear, followed by deeper exploration under ambiguity. Overall, orchestration plays a central role in both preference elicitation and coordination of information throughout the interaction. 

\subsection{Cross-Session Memory}
Cross-session memory improves performance under underspecified queries, increasing semantic alignment and outperforming the no-memory condition. This aligns with prior work on memory-augmented LLM systems\cite{zhong2024memorybank,park2023generative}, where persistent memory enables more personalized and context-aware behavior across interactions.

Its effectiveness, however, depends on how stored preferences interact with current intent. As preferences evolve, previously stored signals may become less informative, leading to misalignment, particularly when outdated or overly generic signals are emphasized. 

Rather than acting as a uniformly beneficial component, memory functions as an additional signal whose impact depends on its integration with interaction context and downstream decisions. When properly aligned, it consistently improves handling of underspecified queries, reinforcing its role in conversational recommendation. 

\subsection{Limitations}

This work relies on a controlled evaluation setting designed to isolate architectural effects. The use of a reduced dataset and simulated interactions enables systematic comparison, but limits representation of real-world variability. Simulated users may not capture the diversity and evolution of real user intent, affecting observed interaction dynamics. Additionally, the limited dataset scale constrains variability in user profiles, and the magnitude of reported gains may differ under larger and more heterogeneous populations. The results should therefore be interpreted as evidence of system-level patterns rather than definitive performance estimates. Extending evaluation to larger-scale and real-world settings remains necessary to assess robustness and generalization.

\section{Conclusion}
\label{sec:conclusion}

This work presents a memory-aware multi-agent architecture for conversational recommendation, integrating LLM-based orchestration, long-term memory, and a hybrid retrieval pipeline. By coordinating preference elicitation, query construction, and recommendation generation, the approach highlights how system-level interactions shape conversational effectiveness.

Across offline, conversational, and cross-session evaluations, the results indicate that orchestration and memory play complementary roles. Supervisory control improves interaction under underspecified intent, while memory enhances personalization when aligned with interaction context. A consistent gap between retrieved and displayed items reveals an exposure bottleneck, indicating that recommendation quality depends not only on retrieval but also on how candidates are presented.

These findings should be interpreted as evidence of system-level effects in a controlled experimental setting, suggesting that effective conversational recommendation requires the joint design of orchestration, memory, and retrieval-ranking components.

Future work includes the development of adaptive memory mechanisms to address preference drift, as well as improved strategies for candidate selection and presentation. In addition, incorporating structured profiling signals represents a promising direction for informing early-stage interaction and improving robustness in cold-start and underspecified scenarios. Extending the evaluation to larger-scale and real-world settings remains essential to assess robustness under more diverse interaction patterns and user behaviors.

\subsubsection{Acknowledgments}

\ifanonym
    This section is omitted due anonyms review.
\else
    This work was supported by Instituto de Ciência e Tecnologia Itaú and Eldorado Research Institute. We acknowledge the use of OpenAI tools for reference discovery and text refinement. Any opinions, findings, conclusions or recommendations expressed in this material are those of the authors and do not necessarily reflect the views of Itaú Unibanco and Instituto de Ciência e Tecnologia Itaú. This document is not and does not constitute or intend to constitute investment advice or any invest-ment service. It is not and should not be deemed to be an offer to purchase or sell, or a solicitation of an offer to purchase or sell, or a recommendation to purchase or sell any securities or other financial instruments. In addition, all data used in this study comply with the Brazilian General Data Protection Law.

    This preprint has not undergone any post-submission improvements or corrections. The Version of Record of this contribution is published in Intelligent Systems, and is available online at https://doi.org/10.1007/978-3-032-39892-5\_25 .
\fi

\bibliographystyle{splncs04}
\bibliography{references}
\end{document}